\documentclass[runningheads]{llncs}
\usepackage[pagebackref]{hyperref}

\usepackage[T1]{fontenc}
\usepackage{graphicx}
\usepackage{color}

\usepackage{array}
\begin{document}
\title{Effect of Gender, Pose and Camera Distance on Human Body 
Dimensions Estimation}
\titlerunning{Effect of Gender, Pose and Camera Distance on HBDE}
%
\author{Yansel González Tejeda\textsuperscript{1}\orcidID{0000-0003-1002-3815} \and
	Helmut A. Mayer\textsuperscript{2}\orcidID{0000-0002-2428-0962}}
\authorrunning{González Tejeda and Mayer}
%
\institute{Department of Artificial Intelligence and Human Interfaces, Paris Lodron University of Salzburg, Austria\\
\textsuperscript{1}\email{yansel.gonzalez-tejeda@stud.sbg.ac.at}\\
\textsuperscript{2}\email{helmut@cs.sbg.ac.at}
}
\maketitle              
\begin{abstract}
Human Body Dimensions Estimation (HBDE) is a task that an intelligent agent can 
perform to attempt to determine human body information from images (2D) or point 
clouds or meshes (3D). More specifically, if we define the HBDE problem as inferring human body measurements from images, then HBDE is a 
difficult, inverse, multi-task regression problem that can be tackled with machine 
learning techniques, particularly convolutional neural networks (CNN). Despite the 
community's tremendous effort to advance human shape analysis, there is a lack of 
systematic experiments to assess CNNs estimation of human body dimensions from 
images. Our contribution lies in assessing a CNN estimation performance in a series of 
controlled experiments. To that end, we augment our recently published neural 
anthropometer dataset by rendering images with different camera distance. We  
evaluate the network inference absolute and relative mean error between the 
estimated and actual HBDs. We train and evaluate the CNN in four scenarios: (1) training with subjects of a specific gender, (2) in a specific pose, (3) sparse camera distance and (4) dense camera distance. Not only our experiments demonstrate that the network can
perform the task successfully, but also reveal a number of relevant facts that contribute to better understand the task of HBDE.

\keywords{Human Body Dimensions Estimation \and Human Body Measurements  \and Deep 
Learning.}
\end{abstract}
\section{Introduction}\label{intro}
Human Body Dimensions Estimation (HBDE) is a task that an intelligent can 
perform to attempt to determine human body information from images (2D) or point 
clouds or meshes (3D). For instance, estimating the height and the shoulder width 
of a person from a picture or a 3D mesh. Being humans in the center of society, one would expect that intelligent agents 
should be able to perceive the shape of a person and reason about it 
from an 
anthropometric perspective, i.e., be capable of accurately estimating her human 
body 
measurements.

This problem can be characterized by specifying the intelligent 
agent's perceptual input. If the HBDE problem is 
circumscribed to inferring human body measurements from images, then HBDE is, 
theoretically, a difficult, inverse, multi-task regression problem.

Practically, HBDE from images is a compelling problem, as well. HBDE plays an 
important role in several areas ranging from digital 
sizing\cite{fitpredictor2021}, 
thought ergonomics\cite{Brito2019}
and computational forensics\cite{thakkar2021feasibility}, to virtual 
try-on\cite{my_coutomized_3D_body_2021} and 
even fashion design and 
intelligent automatic door systems\cite{Kiru2020IntelligentAD}. Moreover, since 
accurately estimating a 
person's body measurements would decrease the probability that the person returns 
clothes acquired online, HBDE has gained attention as an important step toward a more individual-oriented clothes manufacture.

Inverse problems such as HBDE can be tackled with convolutional neural networks 
(CNN). 
However, most studies in the field of HBDE have only focused on investigating to 
what 
extent CNNs can predict body measurements. A number of factors can affect this 
prediction, but researchers have not treated them in depth. What is not yet clear is the impact of the person's gender, pose, and camera 
distance 
with respect to the subject, on the estimation performance.

In this paper, we investigate these dependencies with a series of experiments. Despite the tremendous effort from researchers to attempt to better understand 
HBDE, there is lack of this kind of experiment in the literature. We believe that 
our contribution will shed light on how a CNN estimate 
HBDs.  Upon publication, we will make our code publicly available for research 
purposes.\footnote{Code under \url{https://github.com/neoglez/gpcamdis_hbde}}

\section{The Problem of Human Body Dimensions Estimation}
As stated above, CNNs can be 
employed to approach the HBDE problem. However, supervised learning 
methods demand large amounts of data. Unfortunately, this kind of data is 
extremely difficult to collect. For the network input, 
several persons must be photographed with the same camera under equal lighting 
conditions. Further, in order to study the effect of pose, the subjects must adopt 
several poses; and to study the effect of camera distance, they would have to be 
again photographed. 
The supervision signal is even more challenging and costly: these same subjects 
must be 
accurately measured with identical methods to acquire their body 
dimensions. This is \textbf{the data scarcity problem in HBDE}.

A possible solution is to generate realistic 3D human meshes and 
calculate HBDs from these meshes. But the HBD calculation is by no means a trivial 
task. Properly defining HBDs 
suffers from two issues: inconsistency and uncertainty.   

HBDs definitions differ depending on their intended purpose. To just mention one 
example, health studies measure waist circumference at the midpoint between the inferior margin of the 
last rib and the iliac crest \cite{waist_brasil_2020}. However, while investigating the height of the waist for clothing pattern design, \cite{Gill_Parker_Hayes_Brownbridge_Wren_Panchenko_2014} found seven different waist definitions and \cite{not_all_body2017} directly enunciated that not all body measurements defined by 3D scanning technologies are valid for clothing pattern. This 
multiplicity of definitions complicates consistent conceptualization for machine 
learning.

Furthermore, HBDs are defined based on skeletal joins and/or body landmarks. These 
reference criteria are highly uncertain and depend on the person 
performing the measurement. A single HBD may exhibit important variability due to 
observer or instrument error. Also, researchers and practitioners base their 
analysis on HBD by presenting a figure of a thin subject with the measurements 
depicted by 
segments without further elucidation.
This approach hinders the HBDs calculation reproducibility.

Formally, the HBDE problem has been defined by \cite{gon_mayer_na} as a deep 
regression problem. Given an image $\mathcal{I}$ from a 3D 
human body with HBD $D$, the goal is to return a set $\hat{D}$ of estimated human body 
dimensions, that is

\begin{equation}  
	\hat{D} = \mathcal{M}(\mathcal{I}(D)).
\end{equation}

The dataset is assumed to be drawn from a 
generating distribution and the deep neural network $\mathcal{M}$ minimizes the prediction error.

\section{Related Work}\label{sec:related_work}
Obviously, human body dimensions are determined by human shape. In the field of Human Shape estimation (HSE), shape has been ambiguously presented either as a parametric model 
acting as proxy to a 3D mesh or directly as a triangular mesh. In a community 
effort to be more precise, the task of shape estimation has been currently 
sharper defined as human mesh recovery, estimation or reconstruction. 
Additionally, pose estimation has been established as 
inferring the location of 
skeleton joints, albeit these not being anatomically correct. In the last five 
years, the body of work in these two fields has exploded. Since human mesh and 
pose 
estimation are barely indirectly related to our work, we will not discuss them 
here. In contrast, we focus on end-to-end adults HBDE from images, i.e., the model input 
are 
images of adult subjects and the output are human body measurements.

Undoubtedly, anthropometry has contributed most to human shape analysis.
Important surveys such as CAESAR (1999)\cite{robinette1999caesar}, ANSUR I and II 
(2017)\cite{ansur_1_2_2017} and NHANES(1999-2021)\cite{nhanes2022}, have collected 
HBDs. 
However, they did not take images of the subjects. This makes unclear how 
the 
CNN input could be obtained. Recently, other datasets have been released for 
specific tasks, e.g., \cite{pini2020baracca} propose a dataset with images and 
seven HBDs for estimation in the automotive context.

Of all these compendiums, CAESAR is probably the most convenient data in 
terms of realism. It contains rigorously recorded human body dimensions and 3D 
scans, 
from 
which realistic images could be synthesized. The project costed six million USD 
(see \cite{robinette1999caesar} executive summary). Consequently, this data is 
highly expensive. Alternatively, we employ a generative model derived from real 
humans, capable of producing thousands of 3D meshes from which we can 
calculate and visualize the HBDs.

Certainly, height is the HBD that has been investigated the 
most\cite{kato1998estimation}, \cite{ams.2008.BenAbdelkaderY08}, 
\cite{guan2009unsupervised}, \cite{momeni2012height}
\cite{EstimatingHeights2014}, \cite{Sriharsha2019}, \cite{gunel2019face}, 
\cite{thakkar2021feasibility}, \cite{martynov_garimella_west_2020}.
Very early work\cite{kato1998estimation} investigated the effect of gender and 
inverted pictures when humans estimate height from images. They quantified 
estimation 
performance using Pearson’s Correlation Coefficient and established that the 
estimated 
and ground truth height where highly correlated. This fact has been confirmed recently by \cite{martynov_garimella_west_2020}, which also concluded that humans estimate height inaccurately. Other HBDs have been explored, e.g., waist\cite{Gill_Parker_Hayes_Brownbridge_Wren_Panchenko_2014} but, in general, they have received significantly less attention.

Strongly related to our work are studies using or generating synthetic data and 
calculating or manually collecting HBDs \cite{Dibra.2016a}, \cite{Dibra.2016b} 
\cite{fitme_2019}, \cite{yanlearning2019}, \cite{Yan2020AnthropometricCM}, 
\cite{silhouette_yan_2020}, \cite{vskorvankova2021automatic}. None of these works 
investigated the effect of gender, pose or camera distance 
in the estimation performance. Here, we explore these interactions.

Recently, \cite{linearRe2022HBMBaseline} proposed a baseline for HBDE 
given height and weight. They claimed that linear regression estimates accurately HBDs when the inputs are height and weight. Like we, this method use ground truth 
derived from the SMPL model\cite{Loper.2015}. Despite their input being different 
to ours, we will use this work for comparison.

A \textit{neural anthropometer} (NeuralAnthro) was introduced by 
\cite{gon_mayer_na}. 
The CNN was trained on grayscale synthetic images of moderate complexity, i.e., no 
background, limited 
human poses, and fix camera perspective and distance. In this work, we go further 
and increase the image complexity, making the input more challenging to the intelligent agent conducting HBDE.

\section{Material and Methods}
We now detail the dataset and CNN (model) of the supervised learning approach that 
governs our experiments. 

\begin{figure}
	\includegraphics[width=\textwidth]{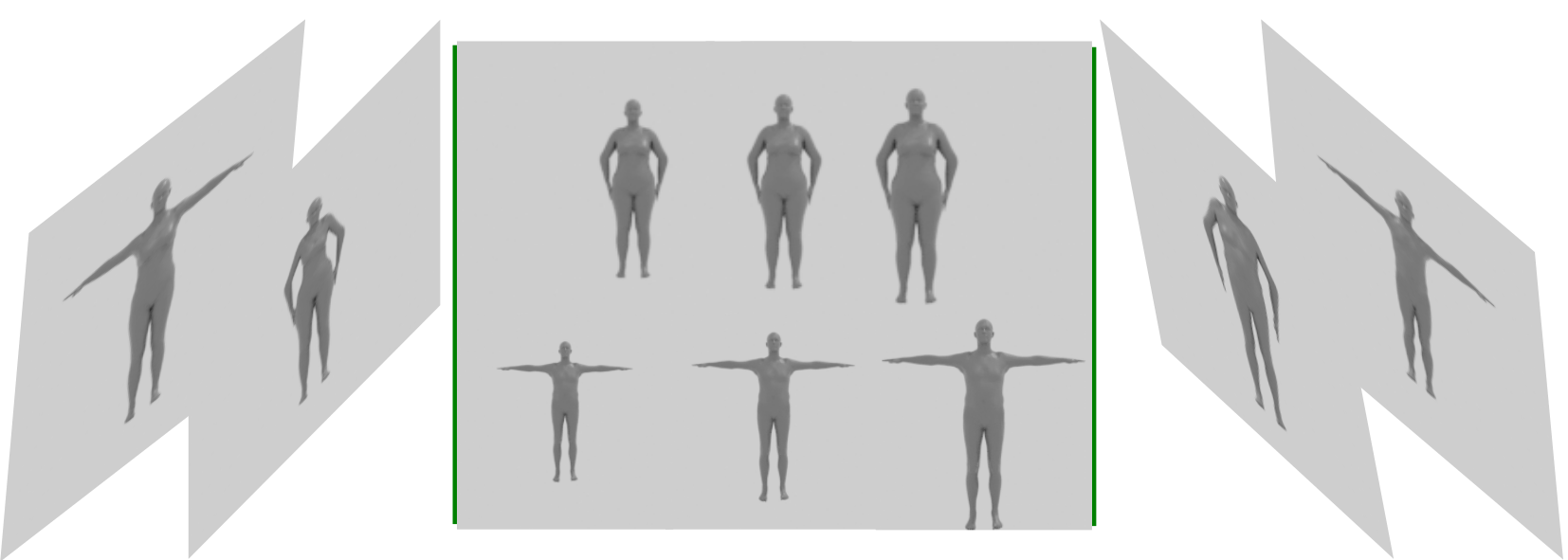}
	\caption{Our curated dataset. We augment the NeuralAnthro dataset, containing 
		images of female (left) and male (right) subjects in pose zero (arms 
		stretched to the sides) and pose one (arms lowered) taken with a camera at 
		a 
		fix distance, by rendering photos with sparse and dense camera distances 
		(center). Note that the subjects appear nearer or farther in the images. 
		All instances are $200\times200$ pixels grayscale  images displaying a 
		single subject.} 
	\label{fig1}
\end{figure}

\subsection{Dataset}
We start with the NeuralAnthro synthetic dataset. The 
reason to use a synthetic dataset is the cost and effort that collecting ”real” 
data would imply. Since coherent pose variability is more difficult to find in 
real datasets, another important aspect is the possibility to vary the subject 
posture to experiment with different poses. While we did not collect our data from 
physical humans, we use the SMPL model, 
which is derived from real humans. SMPL is the most employed model in academia and 
industry for its realism and simplicity \cite{shi2020review}.

\subsubsection{Input}
Figure \ref{fig1} depicts our dataset. We obtained the 3D meshes, $6000$ female 
and 
male 
subjects in pose zero and pose one (total $12 000$ meshes), from the neural 
anthropometer dataset\cite{gon_mayer_na}.

Using the current standard method to 
employ a render engine to produce the mesh corresponding images, we 
simulate the cinematographic technique of \textit{tracking back} (sparsely and 
densely varying the camera distance to the mesh), as follows.

\textbf{Sparse camera distance}: back tracking by placing the camera at 
	distances $4 \; m$, $5 \; m$ and $6 \; m$.
	
\textbf{Dense camera distance}: back tracking by randomly placing the 
	camera between distances $4,2 \; m$ and $7,2 \; m$.

In total, we synthesize $72 000$ pictures from the $12 000$ meshes. The images 
correspond to meshes of a specific gender and a definite pose, taken at specific 
camera distance with respect to the subject. 

\subsubsection{Supervision Signal}
While the data scarcity problem is the major challenge in HBDE, another problem 
is measurement inconsistency. There is no consensus regarding the 
correct manner to define a 
specific measurement, let alone several of them. The problem arises even when HBDs are 
automatically computed by 3D scanning 
technologies\cite{my_coutomized_3D_body_2021}, making manually corrections 
unavoidable. The united method introduced by \cite{gon_mayer_na} with Sharmeam 
(\textbf{Sh}oulder width, right and left 
\textbf{arm}s length and inse\textbf{am}) and Calvis (\textbf{C}hest, 
w\textbf{a}ist and pe\textbf{lvis} circumference plus height) allows us to resolve 
the 
inconsistency issue because it provides a proper method to calculate eight HBDs. 
Additionally, it agrees, to a large extent, 
with 
anthropometry and tailoring.

\subsection{Neural Anthropometer}

The NeuralAnthro is a small, easily deployable CNN that we use to conduct our 
experiments. We use the same experimental setting as in the original 
paper\cite{gon_mayer_na}, i.e., we 
train for $20$ epochs and use mini-batches of size $100$. We report results 
based on 5-fold cross-validation. We minimize the mean squared error between the 
actual and the estimated HBDs using stochastic gradient descent with a momentum 
$0.9$; the learning rate is set to $0.01$.
\section{Results and Discussion}
For the presentation of the results we use the following abbreviations: shoulder 
width (SW), right 
arm length (RAL), left arm length (LAL), inseam (a.k.a. crotch height ) (I), chest 
circumference (CC), 
waist circumference (WC), pelvis circumference (PC) and height (H). Average MAD 
(AMAD) and Average RPE (ARPE) are both represented by a capital A. The figures we 
present are interesting in several ways. Due to space restrictions we can not 
discuss exhaustively all their aspects. Therefore, we examine the most salient 
results.

\begin{figure}
	\includegraphics[width=\textwidth]{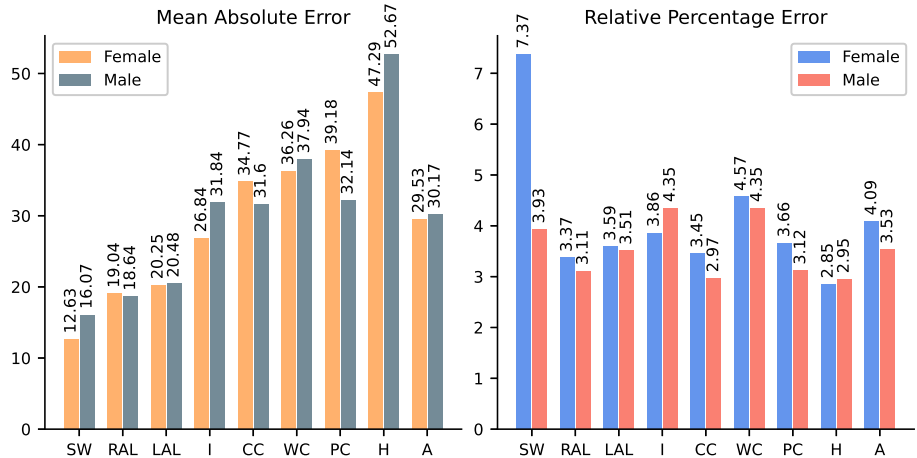}
	\caption{Effect of gender on HBDE. Left: we display Mean 
		Absolute Error (MAE) in $mm$;  right: Relative Percentage Error 
		(RPE). } \label{figOnlyGender}
\end{figure}

\subsection{Effect of Gender}
We start our discussion by evaluating the network performance when the input are 
images from humans of a specific gender in pose zero or one. We define training 
with two gender as 
\textit{unisex training} and \textit{gender training} when the input are subjects 
of a 
specific gender. 
Figure \ref{figOnlyGender} shows the results.

Like \cite{EstimatingHeights2014}, we observe that height estimation is more 
accurate in unisex training, compared to gender training (RPE $1.58$ unisex 
training reported in 
\cite{gon_mayer_na} vs. gender training
$2.85$ female and $2.95$ male).

For the network, it is considerably more difficult to estimate female gender 
training SW than male gender training 
SW. Although female gender training SW MAE is lower than male gender training 
($12.63mm$ vs. $16.07mm$), the inverse relation can be observed, when considering 
RPE ($7.37$ vs. $3.93$).

Curiously, regarding the effect of gender, the CNN and humans appear to estimate height differently. Unlike \cite{kato1998estimation}'s results, Fig.\ref{figOnlyGender} shows that the female height estimation error (RPE $2.85$) is lower than male (RPE $2.95$). Perhaps it is not surprisingly, that this relation holds for inseam as well (RPE $3.86$ vs. $4.35$).
With the exceptions of these two HBDs, the RPE of estimating other HBDs is larger for female as for male subjects.

\subsection{Effect of Pose}

\begin{figure}
	\includegraphics[width=\textwidth]{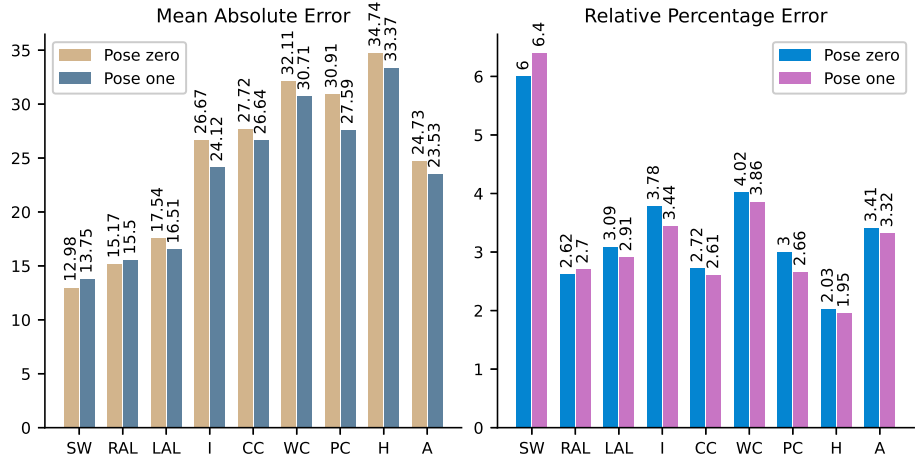}
	\caption{Effect of pose on HBDE. Left: we display Mean 
		Absolute Error (MAE) in $mm$;  right: Relative Percentage Error 
		(RPE). } \label{figOnlyPose}
\end{figure}

Figure \ref{figOnlyPose} presents the breakdown of the estimation error when we 
train the network individually with images of humans in pose zero and pose one 
(\textit{multi-pose training}). Surprisingly, the network estimated shoulder width 
more poorly when the subject was in pose one as in pose zero (RPE $6.4$ vs. 
$6.0$). One would expect that estimating SW would be easier when 
the subject is in pose one, because the arms are lowered, and, therefore, the 
shoulder joints could be easier recognized.

\subsection{Effect of Camera Distance}

The most interesting finding was that the network is able to accurately estimate 
all HBDs independently of the camera distance to the person (ARPE $3.04$, $3.03$, 
$2.96$, $3.57$ and $3.11$), when training with sparse camera distance $4 \; m$, $5 
\; m$ and $6 \; m$ and randomly chosen 
camera distance respectively. This fact 
challenges intuition, e.g., contradicts current research claiming that the network 
can only correctly estimate height if the evaluation is performed for a particular 
camera 
distance\cite{linearRe2022HBMBaseline}. But this finding is in accordance to when 
humans estimating
height as reported in preliminary work\cite{kato1998estimation}.

 \begin{figure}
	\includegraphics[width=\textwidth]{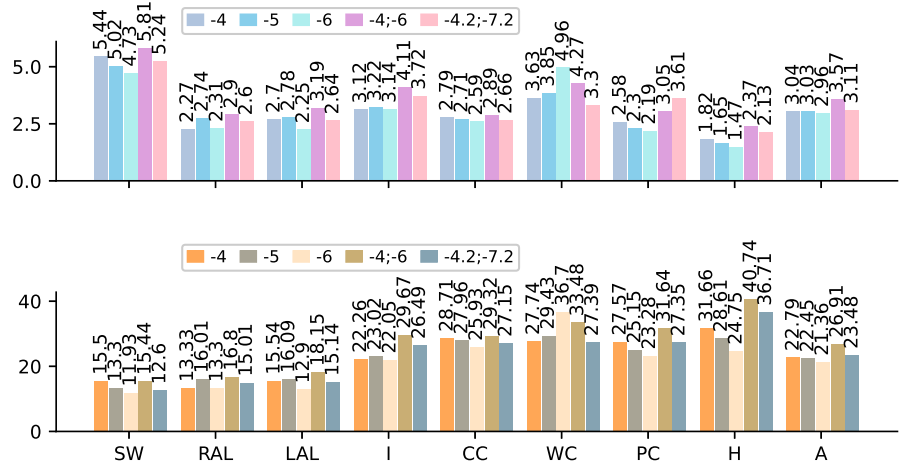}
	\caption{Effect of camera distance on HBDE. We placed the camera at distances 
	$-4 \; m$, $-5 \; m$ and $-6 \; m$, and randomly distances sampled from 
	$-4,2 \; m$ to $-7,2 \; m$ with respect to the subject. Top: Relative 
		Percentage Error (RPE); bottom: Mean Absolute Error (MAE) in $mm$.} 
	\label{figOnlyCamDis}
\end{figure}

\subsection{Quantitative Comparison to Related Work}
Although we did not aim to present a method that outperform SOTA estimation 
methods, we discuss comparative quantitative results for completeness. Basically, 
we compare to 
NeuralAnthro's\cite{gon_mayer_na} original results, the best baseline results 
(Baseline I = 2) on ANSUR data in \cite{linearRe2022HBMBaseline}, a recently 
published study on height estimation from real images by humans\cite{martynov_garimella_west_2020}, and the 
ANSUR II 
allowable error.

We have been eminently cautious in comparing our results in the task 
of human body dimension estimation. Several reasons hinder a fair comparison and 
constitute a major obstacle to advance the field.

First, in the literature, Mean Absolute Error (MAE) and Mean Absolute Difference 
(MAD) refer to the same error quantity. Also, Relative Percentage Error (RPE) 
has not been consistently reported. RPE is important 
because human body dimensions 
are not in the same scale. For 
instance, probably $40mm$ MAE, say, account for a lower height 
estimation error (better performance) as for head circumference error (worst 
performance). Besides MAD and RPE, 
estimation performance has been reported by Mean$\pm$Std. Dev and 
\textit{success rate}\cite{Yan2020AnthropometricCM}, seemingly \textit{Expert 
ratio}. This 
inconsistency in reporting results complicates significantly the comparison with 
other 
research. Second, most method's input are 3D, therefore, inadequate for comparison 
to 2D methods.

\begin{table}
	\centering
	\caption{Comparison to four related methods. We compare estimation performance 
		in 
		terms of MAD error in $mm$. We do not present HBDs that are 
		not comparable (na: not 
		applicable). Minimal errors are bold and we emphasized ANSUR II 
		allowable error. Additionally, we enclosed in parenthesis our 
		experiment setting that achieved best estimation 
		results.}\label{tab:comparison}
	\begin{tabular}{|m{2.7cm}|m{0.88cm}|m{0.88cm}|m{0.88cm}|m{0.88cm}|m{0.88cm}|
			m{2cm}|m{0.88cm}|m{0.88cm}|}
		\hline
		Method & SW & RAL & LAL & I & CC & WC & PC & H\\
		\hline
		Baseline (I = 2) \cite{linearRe2022HBMBaseline} & na & na & na & na & 29.1
		& 37.9 & \textbf{21.6} & na\\[6pt]
		NeuralAnthro \cite{gon_mayer_na} & 12.54 & \textbf{12.98} & 13.48 & 
		\textbf{12.17} & 
		\textbf{25.22} & 27.53 & 25.85 & 27.34\\[6pt]
		Our experiments & \textbf{11.93} ($6m$) & 13.30 ($6m$) & 
		\textbf{12.9} ($6m$) & 22.05 ($6m$) & 25.93 ($6m$) & \textbf{27.39} 
		($-4.2;-7.2$)$m$ & 23.28 
		($6m$) & \textbf{24.75} ($6m$)\\[10pt]
		Humans observing \textbf{real images} \cite{martynov_garimella_west_2020} 
		& na & na & na & na & na 
		& na & na & 64.0\\[10pt]
		Humans, ANSUR (Allowable error ANSUR II)\cite{paquette2009anthropometric} 
		& na & na & na & \textbf{\emph{10.0}} & \textbf{\emph{14.0}} & 
		\textbf{\emph{12.0}} & \textbf{\emph{12.0}} & 
		\textbf{\emph{6.0}}\\
		\hline
	\end{tabular}
\end{table}

Third, we require
that methods' result has been reported persistently in the literature. Neither we 
compare to results reported online 
that are not longer available, nor to results that has been used for comparison 
but we were not able to locate in 
the original cited paper. Last, in the literature, different 
datasets have been used 
for comparing. This might render previous and this comparison 
counterproductive. For example, see ANSUR\cite{paquette2009anthropometric} App. G 
for an extensive account on comparability limitations.

Table \ref{tab:comparison} shows the comparison. The input to Baseline is not 
images (like ours) but height and weight. However, that research does establish 
a 
conceptual baseline: HBDE methods should estimate body 
measurements with higher accuracy compared to regression. This statement should 
not be categorically interpreted. Methods requiring images as input without any 
other information are more challenging and, therefore, might exhibit less 
accuracy. As it can be seen,  NeuralAntro estimates more precisely RAL, I and CC 
as the regression baseline, when applicable, and all of our experiment settings. 
This might happen because NeuralAnthro was trained and evaluated with fixed camera distance. 
The network probably found more difficult learning when trained with three different camera distances. Nevertheless, being SW the most difficult 
HBD to estimate, our experiment with one camera distance at $6m$ manifests the 
best estimation performance. Moreover, our experiment setting with randomly selected camera distances shows 
the best WC estimation performance. 

As the authors indicate in 
\cite{martynov_garimella_west_2020}, height estimation by humans exhibits poor  
performance. The cause is, probably, that the persons estimated the HBD from real 
images, which is the input with highest complexity, compare to synthetic 
controlled data.

Estimation error of all HBD lies over the ANSUR II allowable error, but the fact that the NeuralAnthro is a small CNN could indicate, that by incrementing the size of the network, the estimation performance could be improved as well.

\section{Conclusions and Future Work}

In this paper, we assessed the performance of a neural network employed to estimate human body measurements from images. To that end, we augmented our recently published dataset containing images of female and male subjects in two 
poses, with images of these subjects synthesized using different camera distances 
with respect to the subjects. We trained a CNN with two 
genders, two different poses and sparse and dense camera distances. After training we evaluated the network performance in terms of MAE and RPE.

The CNN estimated HBDs of male subjects more accurately than those of females. The shoulder width predictions exhibit a surprising pose dependency. The width
is estimated more correctly for subjects with arms spread out to the side
(compared to subjects with lowered arms, where the contours of the shoulders
are more pronounced). In contrast to our expectations, network performance
decreases only slightly when perceiving humans from a range of (camera)
distances instead of a fixed distance; given that the person is completely visible in the image. In general, shoulder 
width is the most difficult HBD to estimate.

\subsection{Future Work}

An important question that needs to be answered is why the estimation is, in 
general, highly accurate (errors are reported in mm). Exploring to what extent synthetic data is representative of the real HBDs would contribute to understand this phenomenon. Increasing the level of realism of the images would probably have the strongest effect in HBDE. Also, investigating the minimum amount of data for conducting HBDs with reasonable accuracy, would help determining bounds to collect a plausible real dataset, therefore, alleviating the data scarcity problem.

%
%
%
\bibliographystyle{splncs04}
\bibliography{egbib}
\end{document}